\documentclass{article} % For LaTeX2e
\usepackage{iclr2018_conference,times}
\usepackage{hyperref}
\usepackage{url}
\usepackage{graphicx} % more modern
\usepackage{amsmath}
\usepackage{amssymb}
\usepackage{subcaption}

\DeclareMathOperator{\E}{\mathbb{E}}

\title{Learning how to explain neural networks:\\ PatternNet and PatternAttribution}

\author{Pieter-Jan Kindermans\thanks{Part of this work was done at TU Berlin, part of the work was part of the Google Brain Residency program.} \\
Google Brain\\
\texttt{pikinder@google.com} \\
\And
Kristof T. Sch\"{u}tt \& Maximilian Alber\\
TU Berlin\\
\texttt{\{maximilian.alber,kristof.schuett\}@tu-berlin.de} \\
\And
Klaus-Robert M\"uller\thanks{KRM is also with Korea University and Max Planck Institute for Informatics, Saarbr\"ucken, Germany} \\
TU Berlin\\
\texttt{klaus-robert.mueller@tu-berlin.de} \\
\And
Dumitru Erhan \& Been Kim \\
Google Brain \\
\texttt{\{dumitru,beenkim\}@google.com}
\And
Sven D\"{a}hne\thanks{Sven D\"{a}hne is now at Amazon}\\
TU Berlin \\
\texttt{sven.daehne@tu-berlin.de}
}

\newcommand{\cov}[2]{\textrm{cov}[#1,#2]}

\newcommand{\bx}{\boldsymbol{x}}
\newcommand{\bu}{\boldsymbol{u}}
\newcommand{\br}{\boldsymbol{r}}
\newcommand{\bv}{\boldsymbol{v}}

\newcommand{\bw}{\boldsymbol{w}}
\newcommand{\ba}{\boldsymbol{a}}
\newcommand{\bs}{\boldsymbol{s}}
\newcommand{\bd}{\boldsymbol{d}}

\newcommand{\fig}[1]{ Fig.~\ref{fig:#1}}
\newcommand{\eq}[1]{Eq.~\eqref{eq:#1}}

\iclrfinalcopy % Uncomment for camera-ready version, but NOT for submission.

\begin{document}

\maketitle

\begin{abstract} 
DeConvNet, Guided BackProp, LRP, were invented to better understand deep neural networks. We show that these methods do not produce the theoretically correct explanation for a linear model. Yet they are used on multi-layer networks with millions of parameters. This is a cause for concern since linear models are simple neural networks. We argue that explanation methods for neural nets should work reliably in the limit of  simplicity, the linear models. Based on our analysis of linear models we propose a generalization that yields two explanation techniques (PatternNet and PatternAttribution) that are theoretically sound for linear models and produce improved explanations for deep networks.
\end{abstract}

\section{Introduction}
%Deep learning made a huge impact on a wide variety of applications \citep{Lecun2015,Schmidhuber2015,krizhevsky2012imagenet,mnih2015human,Silver2016,Sutskever2014}.
%Current models have millions of parameters spread over many layers with nonlinear activation functions in between. 
%They learn efficient and powerful representations, but are often considered a `black-box'. In attempts to gain insight into how these models operate, a variety techniques have been proposed~\citep{Simonyan2013,Yosinski2015,Nguyen2016,Baehrens2010,Bach2015,Montavon2017,Zeiler2014,Springenberg2014,Zintgraf2017,Mukund2017,Smilkov2017}.
%A common approach is to explain individual classifier decisions using back-projections to input space. Techniques that make us of this principle include saliency maps from network gradients~\citep{Baehrens2010,Simonyan2013}, DeConvNet \citep[DCN]{Zeiler2014}, Guided BackProp \citep[GBP]{Springenberg2014}, Layer-wise Relevance Propagation \citep[LRP]{Bach2015} and the Deep Taylor Decomposition \citep[DTD]{Montavon2017}.

Deep learning made a huge impact on a wide variety of applications \citep{Lecun2015,Schmidhuber2015,krizhevsky2012imagenet,mnih2015human,Silver2016,Sutskever2014} 
and recent neural network classifiers have become extremely good at detecting relevant \textit{signals} (say, the presence of a cat) contained in input data points such as images by filtering out all other, non-relevant and \textit{distracting} components also present in the data.
This separation of signal and distractors is achieved by passing the input through many layers with millions of parameters and nonlinear activation functions in between until finally at the output layer, these models yield a highly condensed version of the signal, e.g. a single number indicating the probability of a cat being in the image.

While deep neural networks learn efficient and powerful representations, they are often considered a `black-box'. In order to better understand classifier decisions and to gain insight into how these models operate, a variety techniques have been proposed~\citep{Simonyan2013,Yosinski2015,Nguyen2016,Baehrens2010,Bach2015,Montavon2017,Zeiler2014,Springenberg2014,Zintgraf2017,Mukund2017,Smilkov2017}.

The aforementioned methods for explaining classifier decisions operate under the assumption that it is possible to propagate the condensed output signal back through the classifier to arrive at something that shows how the relevant signal was encoded in the input and thereby explains the classifier decision. Simply put, if the classifier detected a cat, the  visualization should point to the cat-relevant aspects of the input image from the perspective of the network.
Techniques that are based on this principle include saliency maps from network gradients~\citep{Baehrens2010,Simonyan2013}, DeConvNet \citep[DCN]{Zeiler2014}, Guided BackProp \citep[GBP]{Springenberg2014}, Layer-wise Relevance Propagation \citep[LRP]{Bach2015} and the Deep Taylor Decomposition \citep[DTD]{Montavon2017}, Integrated Gradients \citep{Mukund2017} and SmoothGrad \citep{Smilkov2017}.

%Methods for explaining classifier decisions have been 
% methodes designed for or in the context of deep neural networks
% in this work, we analyse their performance on simplest neural network: linear model
% we find significant short-comings

The merit of explanation methods is often proven by applying them to state-of-the-art deep learning models in the context of high dimensional real world data, such as ImageNet. Here we begin with a different approach. 
We first take a step back and analyze explanation methods in the context of the simplest neural network setting: a purely linear model and data stemming from a linear generative model. We chose this simplified setup because it allows us to (i) fully control how signal and distractor components are encoded in the input data and (ii) analytically track how the resulting explanation relates to the known signal component. 
This analysis allows us to highlight shortcomings of current explanation approaches that carry over to non-linear models as well. 

On the basis of our findings, we then propose PatternNet and PatternAttribution, which alleviate these flaws. Finally we apply our methods to practically relevant networks and datasets, and show that our approach produces qualitatively improved signal visualizations and attributions (see \fig{fig1} and \fig{fig5}). In addition to the qualitative evaluation, we also experimentally verify whether our proposed theoretical model holds up empirically (see \fig{fig3}).

The remainder of the paper is structured as follows: visualization in linear models is analyzed in section 2. Section 3 relates this analysis to existing approaches for neural network visualization. Section 4 introduces PatternNet and PatternAttribution, which then is evaluated in section 5 before concluding.

\begin{figure*}
\centering
\includegraphics[width=\textwidth]{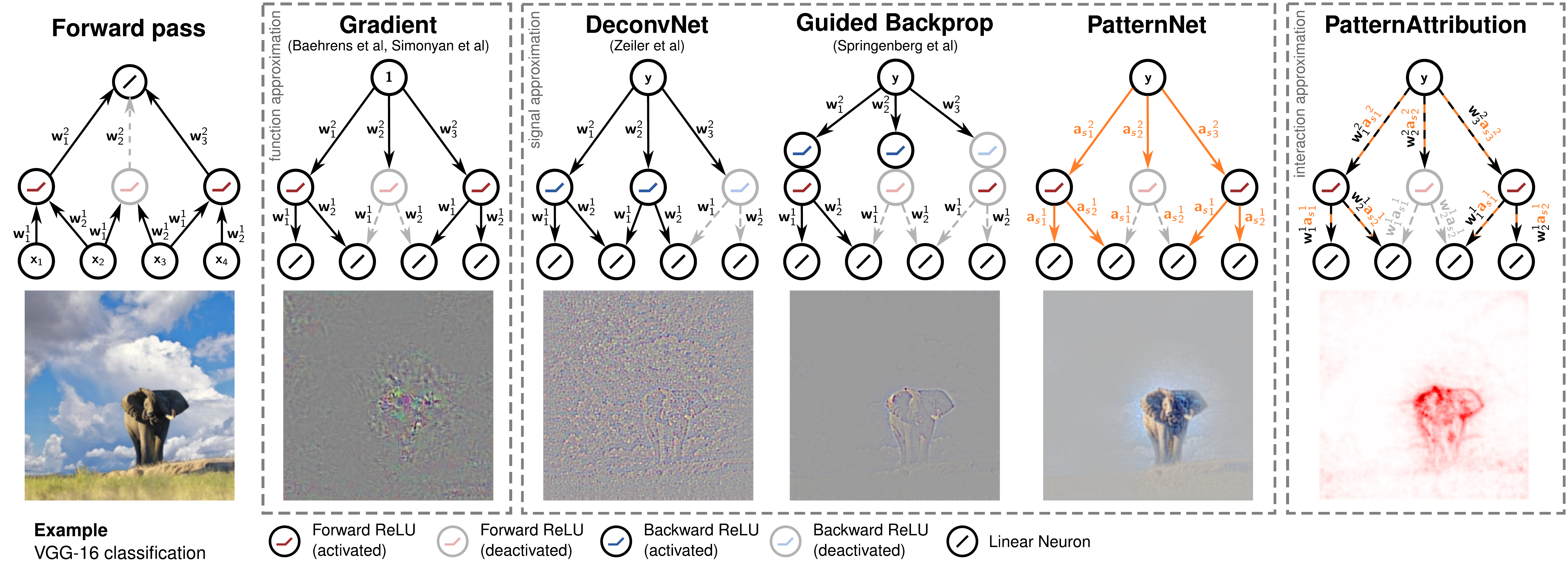}
\caption{Illustration of explanation approaches. Function and signal approximators visualize the explanation using the original color channels. The attribution is visualized as a heat map of pixel-wise contributions to the output \label{fig:fig2}}
\end{figure*}

\paragraph{Notation and scope}
Scalars are lowercase letters ($i$), column vectors are bold ($\bu$), element-wise multiplication is ($\odot$). The covariance between  $\bu$ and $\bv$ is $\cov{\bu}{\bv}$, the covariance of $\bu$ and $i$ is $\cov{\bu}{i}$. The variance of a scalar random variable $i$ is $\sigma^2_{i}$. Estimates of random variables will have a hat ($\hat{\bu}$). 
We analyze neural networks excluding the final soft-max output layer. 
To allow for analytical treatment, we only consider networks with linear neurons optionally followed by a rectified linear unit (ReLU), max-pooling or soft-max. 
We analyze linear neurons and nonlinearities independently such that every neuron has its own weight vector.
These restrictions are similar to those in the saliency map~\citep{Simonyan2013}, DCN~\citep{Zeiler2014}, GBP~\citep{Springenberg2014}, LRP~\citep{Bach2015} and DTD \citep{Montavon2017}.
Without loss of generality, biases are considered constant neurons to enhance clarity.

\section{Understanding linear models}
\label{sec:linearmodels}
\begin{figure*}
\centering
\includegraphics[width=\textwidth]{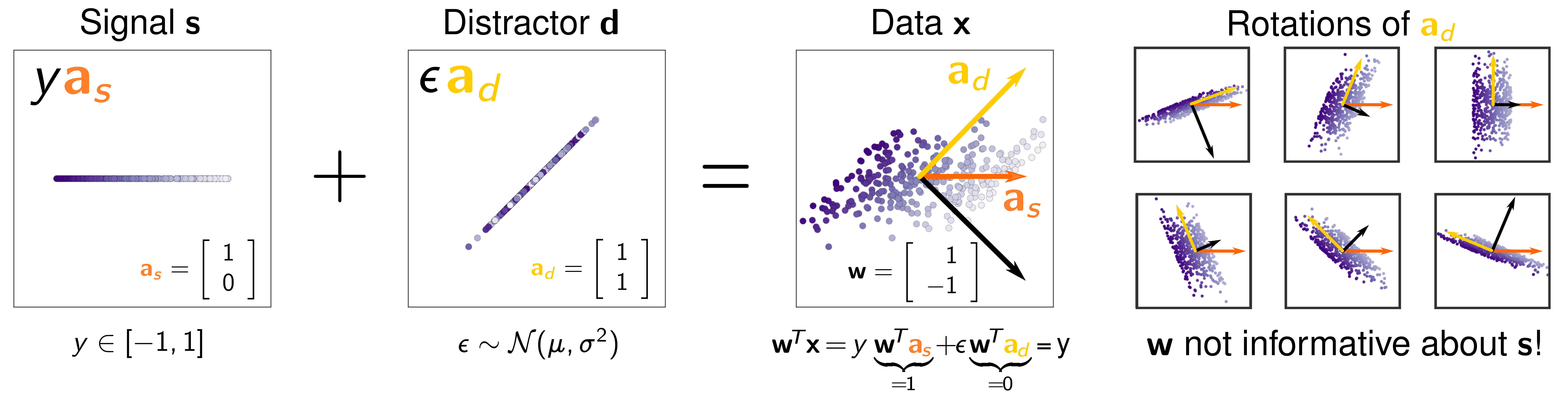}
\caption{
For linear models, i.e., a simple neural network, the weight vector does \textbf{not} explain the signal it detects \cite{Haufe2014}. The data $\bx = y \ba_s + \epsilon \ba_d$ is color-coded w.r.t. the output $y=\bw^T\bx$. Only the signal $s = y \ba_s$ contributes to $y$. The weight vector $\bw$  does not agree with the signal direction, since its primary objective is canceling the distractor. Therefore, rotations of the basis vector $\ba_d$ of the distractor with constant signal $\bs$ lead to rotations of the weight vector (\textbf{right}).
 \label{fig:fig1}}
\end{figure*}

Before moving to deep networks, we analyze the behavior of a linear model (see \fig{fig1}).% in a controlled setting.
Consider the following toy example where we generate data $\bx$ as:
\begin{align*}
\bx &= \bs + \bd &
 \bs &=  \ba_s y, &\textrm{with } \ba_s       & =\left(1,0\right)^T, \,~~~~~~~ y \in \left[-1,1\right] \\
& & \bd &= \ba_d \epsilon, &\textrm{with } \ba_d & =\left(1,1\right)^T, \,~~~~~~~ \epsilon \sim \mathcal{N}\left(\mu,\sigma^2\right).
\end{align*}
We train a linear regression model to extract $y$ from $\bx$.
By construction, $\bs$ is the {\em signal} in our data, i.e., the part of $\bx$ containing information about $y$. 
Using the terminology of \citet{Haufe2014} %and \citet{Zeiler2014}
%\note{max to kristof: why did you add Zeiler here?}, the direction $\ba_s$ along which the signal varies is called the input {\em pattern}.
the {\em distractor} $\bd$ obfuscates the signal making the detection task more difficult.
To optimally extract $y$, our model has to be able to filter out the distractor $\bd$.
This is why the weight vector is also called the {\em filter}.
In the example, $\bw=\left[1,-1\right]^T$ fulfills this task.

From this simple example, we can make several observations:
The optimal weight vector $\bw$ does \emph{not} align, in general, with the signal direction $\ba_s$, but tries to filter the contribution of the distractor~(see \fig{fig1}).
This optimally solved when the weight vector is orthogonal to the distractor $\bw^T \bd = 0$.
Therefore, when the direction of the distractor $\ba_d$ changes, $\bw$ must follow, as illustrated on the right hand side of the figure.
On the other hand, a change in signal direction $\ba_s$ can be compensated for by a change in sign and magnitude of $\bw$ such that $\bw^T\ba_s=1$, but the direction stays constant.

This implies that in the situation where we have a signal and distractor(s), the direction of the weight vector in a linear model is largely determined by the distractor. This is essential to understand how linear models operate. This also indicates that given only the weight vector, we cannot know what part of the input produces the output $y$. This is the direction $\ba_s$ and must be learned from data. Please note that the linear problem above is convex, therefore a weight vector obtained by optimizing the model would converge to the analytical solution defined above.

Now assume that we have no distractor but instead we have additive isotropic Gaussian noise.
It is easy to verify that the mean of the noise is of little importance since it can be compensated for with a bias change. Therefore, we only have to consider case where the noise is zero mean. Because isotropic Gaussian noise does not contain any correlations or structure, the only way to cancel it out is by averaging over different measurements. It is \emph{not} possible to cancel it out effectively by using a well-chosen weight vector. However, it is well known that adding Gaussian noise shrinks the weight vector and corresponds to l2 regularization. In the absence of a structured distractor, the smallest weight vector $\bw$ such that $\bw^T\ba_s=1$ is the one in the direction of the signal. Therefore in practice both these effects influence the actual weight vector. 

Considering the reasoning above, we have to wonder under which conditions we are working in a deep neural network. Especially since DeConvNet and Guided BackProp produce crisp visualizations using (modified) gradients. For this reason we will perform the following quantitative and qualitative experiments which indicate that our theory also holds for a deep network: % kristof: too long. don't spoil the later sections
\begin{itemize}
\item In \fig{fig3} we have evaluated how well the weight vector or a learned direction captures the information content in the input of every single neuron in VGG16. This experiment empirically shows that a learned direction captures more information than the direction defined by the weight vector. This indicates that we are working (largely) in the distractor-regime.
\item This experiment is confirmed by an image degradation experiment in \fig{fig4}.
\item It is also corroborated by the qualitative inspection of the visualizations in \fig{fig2}, \fig{fig5} and \fig{fig6}.
\end{itemize}
Finally, there is also an intuitive argument. Neural networks are considered layer-wise feature extractors that add more invariances as we move through the layers. Since cancelling out a distractor is adding an invariance, the proposed theory fits this interpretation well.

Before moving on to the discussion of interpretability methods, we would like to remind the reader of the terminology that is used throughout this manuscript:
The {\em filter} $\bw$ tells us how to extract the output $y$ optimally from data $\bx$.
The \emph{pattern} $\ba_s$ is the direction in the data along which the desired output $y$ varies.
Both constitute the \emph{signal} $\bs=\ba_s y$, i.e., the contributing part of $\bx$. 
The {\em distractor} $\bd$ is the component of the data that does not contain information about the desired output.

\section{Overview of explanation approaches and their behavior}
\label{sec:methods}
In this section, we take a look at a subset of explanation methods for individual classifier decisions and discuss how they are connected to our analysis of linear models in the previous section. 
\fig{fig2} gives an overview of the different types of explanation methods which can be divided into function, signal and attribution visualizations. 
These three groups all present different information about the network and complement each other.

\paragraph{Functions --  gradients, saliency map } %\note{kristof: Should we just call this filters instead of functions?}}
Explaining the function in input space corresponds to describing the operations the model uses to extract $y$ from $\bx$.
Since deep neural networks are highly nonlinear, this can only be approximated. The saliency map estimates how moving along a particular direction in input space influences $y$ (i.e., sensitivity analysis) where the direction is given by the model gradient \citep{Baehrens2010,Simonyan2013}.
In case of a linear model $y = \bw^T \bx$, the saliency map reduces to analyzing the weights $\partial y / \partial \bx = \bw$.
Since it is mostly determined by the distractor, as demonstrated above%
 %and verified empirically in \fig{fig3}  %kristof: we first need to explain how evaluation works
 , it is not representing the signal. It tells us how to extract the signal, not what the signal is in a deep neural network. 

\paragraph{Signal -- DeConvNet, Guided BackProp, PatternNet} The signal $\bs$ detected by the neural network is the component of the data that caused the networks activations.
\citet{Zeiler2014} formulated the goal of these methods as "[...] to map these activities back to the input pixel space, showing what input pattern originally caused a given activation in the feature maps".

In a linear model, the signal corresponds to $\bs=\ba_s y$.
The pattern $\ba_s$ contains the signal direction, i.e., it tells us where a change of the output variable is expected to be measurable in the input \citep{Haufe2014}.
Attempts to visualize the signal for deep neural networks were made using DeConvNet \citep{Zeiler2014} and Guided BackProp \citep{Springenberg2014}. 
These use the same algorithm as the saliency map, but treat the rectifiers differently (see \fig{fig2}):
DeConvNet leaves out the rectifiers from the forward pass, but adds additional ReLUs after each deconvolution, while Guided BackProp uses the ReLUs from the forward pass as well as additional ones.
The back-projections for the linear components of the network correspond to a superposition of what are assumed to be the signal directions of each neuron.
For this reason, these projections must be seen as an approximation of the features that activated the higher layer neuron.
It is not a reconstruction in input space \citep{Zeiler2014}.

For the simplest of neural networks -- the linear model -- these visualizations reduce to the gradient\footnote{In tensorflow terminoloy: linear model on MNIST can be seen as a convolutional neural network with  VALID padding and a 28 by 28 filter size. }. %\note{kristof: perhaps note that a linear model can be seen as a convnet with a large filter? otherwise the connection from deconvolution to fully-connected is not here}
They show the filter $\bw$ and \emph{neither} the pattern $\ba_s$, \emph{nor} the signal $\bs$. 
Hence, DeConvNet and Guided BackProp do not guarantee to produce the detected signal for a linear model, which is proven by our toy example in \fig{fig1}.
Since they do produce compelling visualizations, we will later investigate whether the direction of the filter $\bw$ coincides with the direction of the signal $\bs$.
We will show that this is \emph{not} the case and propose a new approach, PatternNet (see \fig{fig2}), to estimate the correct direction that improves upon the DeConvNet and Guided BackProp visualizations.

\paragraph{Attribution -- LRP, Deep Taylor Decomposition, PatternAttribution}
Finally, we can look at how much the signal dimensions contribute to the output through the layers. 
This will be referred to as the \emph{attribution}.
For a linear model, the optimal attribution would be obtained by element-wise multiplying the signal with the weight vector:
$
\br^{input} = \bw \odot \ba y,
$
with $\odot$ the element-wise multiplication. % Note odot was not definied...
\cite{Bach2015} introduced \emph{layer-wise relevance propagation} (LRP) as a decomposition of pixel-wise contributions (called \emph{relevances}).
 \cite{Montavon2017} extended this idea and proposed the deep Taylor decomposition (DTD).
The key idea of DTD is to decompose the activation of a neuron in terms of contributions from its inputs.
This is achieved using a first-order Taylor expansion around a root point $\bx_0$ with $\bw^T\bx_0=0$.
The relevance of the selected output neuron $i$ is initialized with its output from the forward pass. The relevance from neuron $i$ in layer $l$ is re-distributed towards its input as: 
 \[
r^{output}_i=y,~~~~~~~~~~~~r^{output}_{j\neq i} = 0,~~~~~~~~~~~~\br^{l-1}_i=\frac{\bw\odot\left(\bx-\bx_0\right)}{\bw^T\bx}r^l_i.
 \]
Here we can safely assume that $\bw^T\bx>0$ because a non-active ReLU unit from the forward pass stops the re-distribution in the backward pass.
This is identical to how a ReLU stops the propagation of the gradient.
The difficulty in the application of the deep Taylor decomposition is the choice of the root point $\bx_0$, for which many options are available.
It is important to recognize at this point that selecting a root point for the DTD corresponds to estimating the distractor $\bx_0 = \bd$ and, by that, the signal $\hat{\bs}=\bx-\bx_0$. PatternAttribution is a DTD extension that learns from data how to set the root point.

Summarizing, the \textbf{function} extracts the \textbf{signal} from the data by removing the distractor.
The \textbf{attribution} of output values to input dimensions shows how much an individual component of the signal contributes to the output, which is what LRP calls \emph{relevance}.

\section{Learning to estimate the signal}
Visualizing the function has proven to be straightforward \citep{Baehrens2010,Simonyan2013}.
In contrast, visualizing the signal \citep{Haufe2014,Zeiler2014,Springenberg2014} and the attribution \citep{Bach2015,Montavon2017,Mukund2017} is more difficult.
It requires a good estimate of what is the signal and what is the distractor. 
In the following section we first propose a quality measure for neuron-wise signal estimators. 
This allows us to evaluate existing approaches and, finally, derive signal estimators that optimize this criterion. 
These estimators will then be used to explain the signal (PatternNet) and the attribution (PatternAttribution). 
All mentioned  techniques as well as our proposed signal estimators treat neurons independently, i.e., the full explanation will be a superposition of neuron-wise explanations.

\subsection{Quality criterion for signal estimators}\label{sec:measure}
Recall that the input data $\bx$ comprises both signal and distractor: $\bx = \bs + \bd, $ and that the signal contributes to the output but the distractor does not. 
Assuming the filter $\bw$ has been trained sufficiently well to extract $y$, we have
\[
\bw^T \bx = y,~~~~~\bw^T \bs= y,~~~~~~\bw^T \bd = 0.
\]
Note that estimating the signal based on these conditions alone is an ill-posed problem. 
We could limit ourselves to linear estimators of the form $\hat{\bs} = \bu(\bw^T\bu)^{-1} y$, with $\bu$ a random vector such that $\bw^T\bu\neq0$.
For such an estimator, the signal estimate $\hat{\bs}=\bu\left(\bw^T\bu\right)^{-1} y$ satisfies $\bw^T \hat{\bs}= y$.
This implies the existence of an infinite number of possible rules for the DTD as well as infinitely many back-projections for the DeConvNet family.

To alleviate this issue, we introduce the following quality measure $\rho$ for a signal estimator $S(\bx)=\hat{\bs}$ that will be written with explicit variances and covariances using the shorthands $\hat{\bd}=\bx-S(\bx)$ and $y=\bw^T\bx$:
\begin{equation}
\rho(S) = 1 - \max_{\bv} corr\left(\bw^T\bx,\bv^T\left(\bx-S({\bx})\right)\right) = 1 - \max_{\bv} \frac{\bv^T\cov{\hat{\bd}}{y}}{\sqrt{\sigma^2_{\bv^T\hat{\bd}} \sigma^2_{y}}}. \label{eq:corr}
\end{equation}
This criterion introduces an additional constraint by measuring how much information about $y$ can be reconstructed from the residuals $\bx - \hat{\bs}$ using a linear projection.
The best signal estimators remove most of the information in the residuals and thus yield large $\rho(S)$.
Since the correlation is invariant to scaling, we constrain $\bv^T\hat{\bd}$ to have variance $\sigma^2_{\bv^T\hat{\bd}} =\sigma^2_{y}$.
Finding the optimal $\bv$ for a fixed $S(\bx)$ amounts to a least-squares regression from $\hat{\bd}$ to $y$. This enables us to assess the quality of signal estimators efficiently. 

\subsection{Existing Signal Estimators}
Let us now discuss two signal estimators that have been used in previous approaches. 

\paragraph{$S_{\bx}$ -- the identity estimator}
The naive approach to signal estimation is to assume the entire data is signal and there are no distractors:
\[
S_{\bx}(\bx)=\bx.
\]
With this being plugged into the deep Taylor framework, we obtain the $z$-rule \citep{Montavon2017} which is equivalent to LRP \citep{Bach2015}. 
For a linear model, this corresponds to 
$\br = \bw \odot \bx$ as the attribution. 
It can be shown that for ReLU and max-pooling networks, the $z$-rule reduces to the element-wise multiplication of the input and the saliency map~\citep{LRPGRAD16,Kindermans2016}. % No max, this is about who proved this not about the saliency map .... \note{Max: shouldn't saliency map be the gradient?}. 
This means that for a whole network, the assumed signal is simply the original input image. 
It also implies that, if there are distractors present in the data, they are included in the attribution:
\[
\br = \bw \odot \bx = \bw \odot \bs + \bw \odot \bd.
\]
When moving through the layers by applying the filters $\bw$ during the forward pass, the contributions from the distractor $\bd$ are cancelled out. However, they cannot be cancelled in the backward pass by the element-wise multiplication. The distractor contributions $\bw \odot \bd$ that are included in the LRP explanation cause the noisy nature of the visualizations based on the $z$-rule. 

\paragraph{$S_{\bw}$ -- the filter based estimator}
The implicit assumption made by DeConvNet and Guided BackProp is that the detected signal varies in the direction of the weight vector $\bw$. 
This weight vector has to be normalized in order to be a valid signal estimator. 
In the deep Taylor decomposition framework this corresponds to the $\bw^2$-rule and results in the following signal estimator:
\[
S_{\bw}(\bx) = \frac{\bw}{\bw^T\bw}\bw^T\bx.
\]
For a linear model, this produces an attribution of the form $\frac{\bw \odot \bw}{\bw^T\bw}y$.
This estimator does not reconstruct the proper signal in the toy example of section \ref{sec:linearmodels}. Empirically it is also sub-optimal in our experiment in \fig{fig3}.

\subsection{PatternNet and PatternAttribution}
We suggest to learn the signal estimator $S$ from data by optimizing the previously established criterion.
A signal estimator $S$ is optimal with respect to Eq.~\eqref{eq:corr} if the correlation is zero for all possible $\bv$: $\forall \bv, \cov{y}{\hat{\bd}}\bv = \boldsymbol{0}$. This is the case when there is no covariance between $y$ and $\hat{\bd}$. Because of linearity of the covariance and since $\hat{\bd}=\bx-S(\bx)$ the above condition leads to
\begin{equation}
\cov{y}{\hat{\bd}}=\boldsymbol{0} \Rightarrow \cov{\bx}{y}=\cov{S(\bx)}{y}.\label{eq:cov_equal}
\end{equation}
It is important to recognize that the covariance is a summarizing statistic and consequently the problem can still be solved in multiple ways. 
We will present two possible solutions to this problem.
Note that when optimizing the estimator, the contribution from the bias neuron will be considered $0$ since it does not covary with the output $y$.

\paragraph{$S_{\ba}$ -- The linear estimator}
A linear neuron can only extract linear signals $\bs$ from its input $\bx$.
Therefore, we could assume a linear dependency between $\bs$ and $y$, yielding a signal estimator:
\begin{equation}
S_{\ba}(\bx)={\ba}\bw^T\bx.\label{eq:linear_estimator}
\end{equation}
Plugging this into \eq{cov_equal} and optimising for ${\ba}$ yields
\begin{equation}
\notag 
\cov{\bx}{y}=\cov{{\ba} \bw^T\bx}{y}={\ba} \cov{y}{y} \Rightarrow
{\ba} =\frac{\cov{\bx}{y}}{\sigma^2_y}.\label{eq:closed_a}
\end{equation}
Note that this solution is equivalent to the approach commonly used in neuro-imaging~\citep{Haufe2014} despite different derivation.
With this approach we can recover the signal of our toy example in section \ref{sec:linearmodels}.
It is equivalent to the filter-based approach only if the distractors are orthogonal to the signal.
We found that the linear estimator works well for the convolutional layers. However, when using this signal estimator with ReLUs in the dense layers, there is still a considerable correlation left in the distractor component (see \fig{fig3}).

\paragraph{$S_{\ba_{+-}}$ -- The two-component estimator}
To move beyond the linear signal estimator, it is crucial to understand how the rectifier influences the training.
Since the gate of the ReLU closes for negative activations, the weights only need to filter the distractor component of neurons with $y > 0$.
Since this allows the neural network to apply filters locally, we cannot assume a global distractor component.
We rather need to distinguish between the positive and negative regime:
\[
\bx = 
\begin{cases}
\bs_+ + \bd_+ & \text{if } y>0\\
\bs_- + \bd_- & \text{otherwise}
\end{cases}
\]
Even though signal and distractor of the negative regime are canceled by the following ReLU, we still need to make this distinction in order to approximate the signal.
Otherwise, information about whether a neuron fired would be retained in the distractor.
Thus, we propose the two-component signal estimator:
\begin{equation}\label{eq:twocomponent}
S_{\ba+-}(\bx)=\begin{cases}
\ba_+\bw^T\bx,~~~~\textrm{if}~\bw^T\bx>0\\
\ba_-\bw^T\bx,~~~~\textrm{otherwise}
\end{cases}
\end{equation}

Next, we derive expressions for the patterns $\ba_+$ and $\ba_-$.
We denote expectations over $\bx$ within the positive and negative regime with $\E_+\left[\bx\right]$ and $\E_-\left[\bx\right]$, respectively. 
Let $\pi_+$ be the expected ratio of inputs $\bx$ with $\bw^T\bx>0$.
The covariance of data/signal and output become:
\begin{align}
\cov{\bx}{y}&=&\pi_+\left(\E_+ \left[\bx y\right]-\E_+ \left[ \bx \right]\E \left[y\right]\right)
		    &~~+&\left(1-\pi_+\right)\left(\E_- \left[\bx y\right]-\E_- \left[ \bx \right]\E \left[y\right]\right)\\
\cov{\bs}{y}&=&\pi_+\left(\E_+ \left[\bs y\right]-\E_+ \left[ \bs \right]\E \left[y\right]\right)
		    &~~+&\left(1-\pi_+\right)\left(\E_- \left[\bs y\right]-\E_- \left[ \bs \right]\E \left[y\right]\right)
\end{align}
Assuming both covariances are equal, we can treat the positive and negative regime separately using Eq.~\eqref{eq:cov_equal} to optimize the signal estimator:
\begin{eqnarray*}
\E_+ \left[\bx y\right]-\E_+ \left[ \bx \right]\E \left[y\right]&=&\E_+ \left[\bs y\right]-\E_+ \left[ \bs \right]\E \left[y\right]
\end{eqnarray*}
Plugging in Eq.~\eqref{eq:twocomponent} and solving for $\ba_+$ yields the required parameter ($\ba_-$ analogous).
\begin{eqnarray}
\ba_+&=&\frac{\E_+ \left[\bx y\right]-\E_+ \left[ \bx \right]\E \left[y\right]}{\bw^T\E_+ \left[\bx y\right]-\bw^T\E_+ \left[ \bx \right]\E \left[y\right]}\label{eq:closed_a+-}
\end{eqnarray}
The solution for $S_{\ba+-}$ reduces to the linear estimator when the relation between input and output is linear. Therefore, it solves our introductory linear example correctly.

\paragraph{PatternNet and PatternAttribution}  % kristof: i think we don't need this anymore since we have the section heading
Based on the presented analysis, we propose PatternNet and PatternAttribution as illustrated in \fig{fig2}.
{\em PatternNet} yields a layer-wise back-projection of the estimated signal to input space. 
The signal estimator is approximated as a superposition of neuron-wise, nonlinear signal estimators $S_{\ba+-}$ in each layer.  It is equal to the computation of the gradient where during the backward pass the weights of the network are replaced by the informative directions. 
In \fig{fig2}, a visual improvement over DeConvNet and Guided Backprop is apparent.

{\em PatternAttribution} exposes the attribution $\bw \odot \ba_+$ and improves upon the layer-wise relevance propagation (LRP) framework~\citep{Bach2015}.
It can be seen as a root point estimator for the Deep-Taylor Decomposition (DTD).
Here, the explanation consists of neuron-wise contributions of the estimated \emph{signal} to the classification score. By ignoring the distractor, PatternAttribution can reduce the noise and produces much clearer heat maps. By working out the back-projection steps in the Deep-Taylor Decomposition with the proposed root point selection method, it becomes obvious that PatternAttribution is also analogous to the backpropagation operation. In this case, the weights are replaced during the backward pass by $\bw\odot\ba_+$.

\section{Experiments and discussion}
To evaluate the quality of the explanations, we focus on the task of image classification. Nevertheless, our method is not restricted to networks operating on image inputs.  We used Theano \citep{Bergstra2010} and Lasagne \citep{Dieleman2015} for our implementation. We restrict the analysis to the well-known ImageNet dataset \citep{Imagenet2015} using the pre-trained VGG-16 model \citep{Simonyan2014}. Images were rescaled and cropped to 224x224 pixels. The signal estimators are trained on the first half of the training dataset. 

The vector $\bv$, used to measure the quality of the signal estimator $\rho(\bx)$ in \eq{corr}, is optimized on the second half of the training dataset. This enables us to test the signal estimators for generalization.  All the results presented here were obtained using the official validation set of 50000 samples.
The validation set was not used for training the signal estimators, nor for training the vector $\bv$ to measure the quality. Consequently our results are obtained on previously unseen data.

The linear and the two component signal estimators are obtained by solving their respective closed form solutions (\eq{closed_a} and \eq{closed_a+-}). With a highly parallelized implementation using 4 GPUs this could be done in 3-4 hours. This can be considered reasonable given that several days are required to train the actual network.
The quality of a signal estimator is assessed with \eq{corr}. Solving it with the closed form solution is computationally prohibitive since it must be repeated for every single weight vector in the network. Therefore we optimize the equivalent least-squares problem using stochastic mini-batch gradient descent with ADAM~\cite{Kingma2014} until convergence. This was implemented on a NVIDIA Tesla K40 and took about 24 hours per optimized signal estimator.

After learning to explain, individual explanations are computationally cheap since they can be implemented as a back-propagation pass with a modified weight vector. As a result, our method produces explanations at least as fast as the work by \cite{Dabkowski2017} on real time saliency. However, our method has the advantage that it is not only applicable to image models but is a generalization of the theory commonly used in neuroimaging~\cite{Haufe2014}.

\begin{figure}
\centering
\includegraphics[trim={5cm 0cm 0cm 0cm},clip,width=\textwidth]{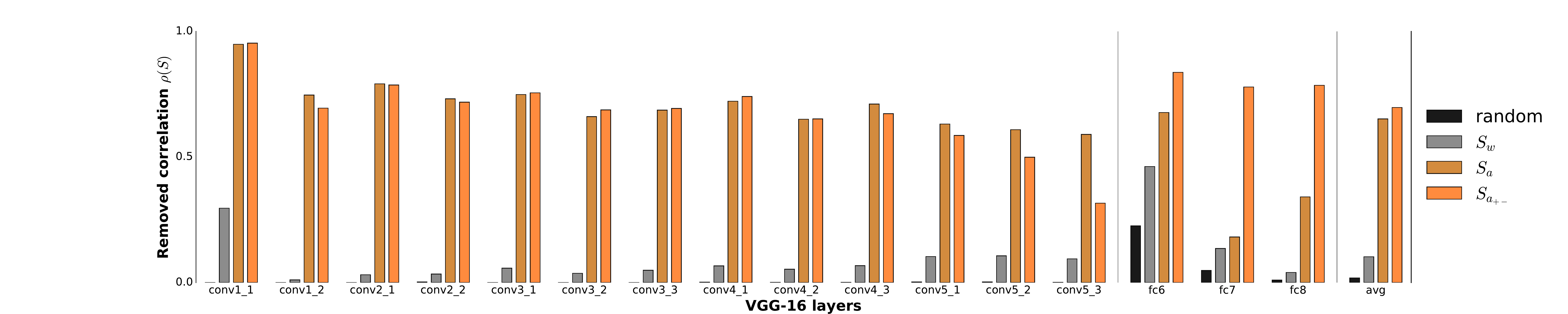}
\caption{Evaluating $\rho(S)$ for VGG-16 on ImageNet. Higher values are better. The gradient ($S_{\bw}$), linear estimator ($S_{\ba}$) and nonlinear estimator ($S_{\ba_{+-}}$) are compared. An estimator using random directions is the baseline. The network has 5 blocks with 2/3 convolutional layers and 1 max-pooling layer each, followed by 3 dense layers. \label{fig:fig3}}
\end{figure}

\begin{figure*}
\begin{subfigure}[b]{0.48\textwidth}
\includegraphics[trim={1cm 0cm 1.5cm 1cm},clip,width=\textwidth]{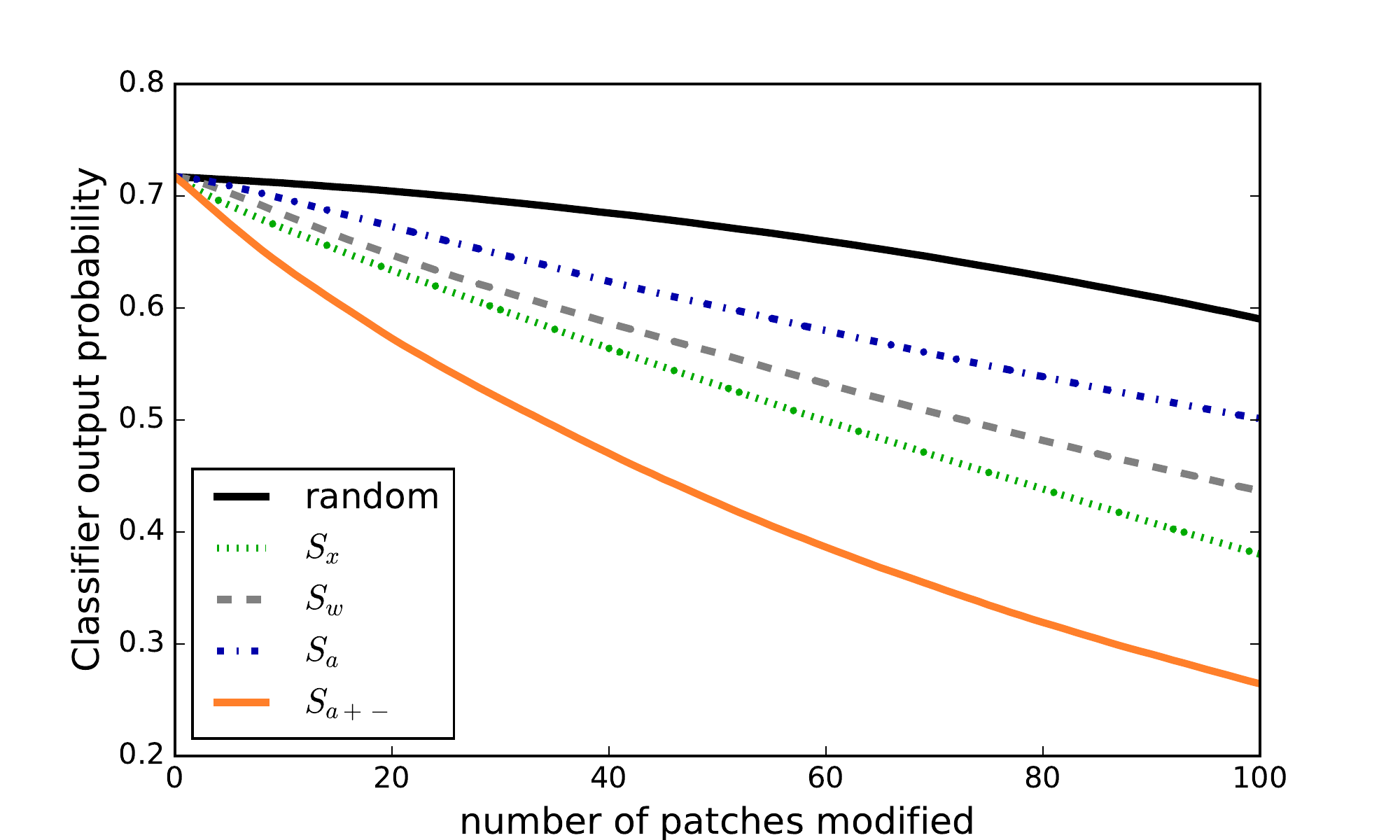} % [trim={1cm 0cm 1.5cm 1cm},clip,width=.33\textwidth]{
\caption{Image degradation experiment on all 50.000 images in the ImageNet validation set. The effect on the classifier output is measured. A steeper decrease is better. \label{fig:fig4}}
\end{subfigure}
~
\begin{subfigure}[b]{0.48\textwidth}
\includegraphics[trim={4cm 13cm 6cm 14cm},clip,width=\textwidth]{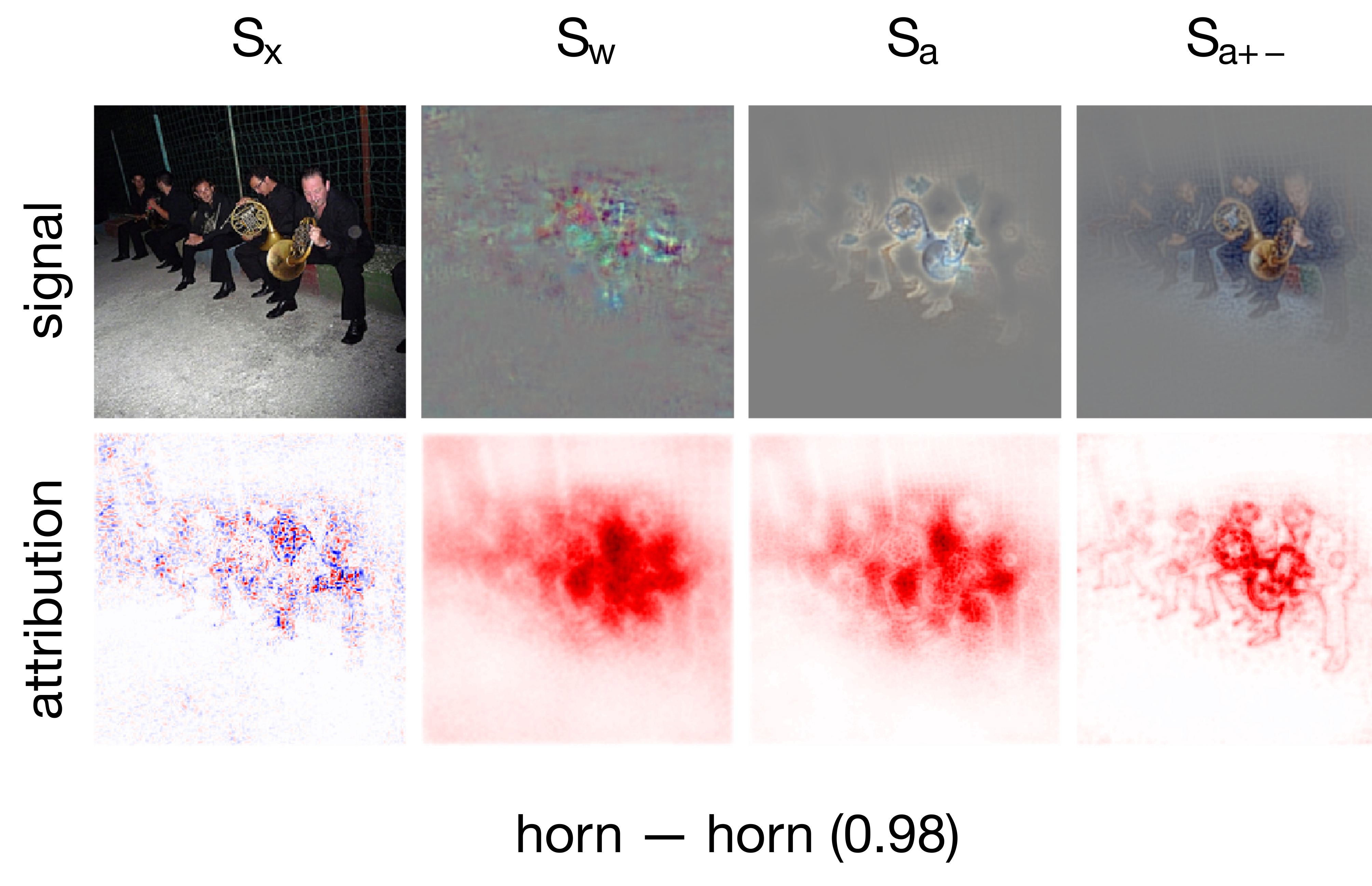} %trim={4cm 13cm 6cm 14cm},clip,width=.33\textwidth
\caption{Top: signal. Bottom: attribution. For the trivial estimator $S_x$ the original input is the signal. This is not informative w.r.t. how the network operates. }
\label{fig:fig5}
\end{subfigure}
\end{figure*}

\begin{figure*}% left, bottom, right,top
\centering
\includegraphics[trim={8cm 24cm 14cm 15cm},clip,width=.96\textwidth]{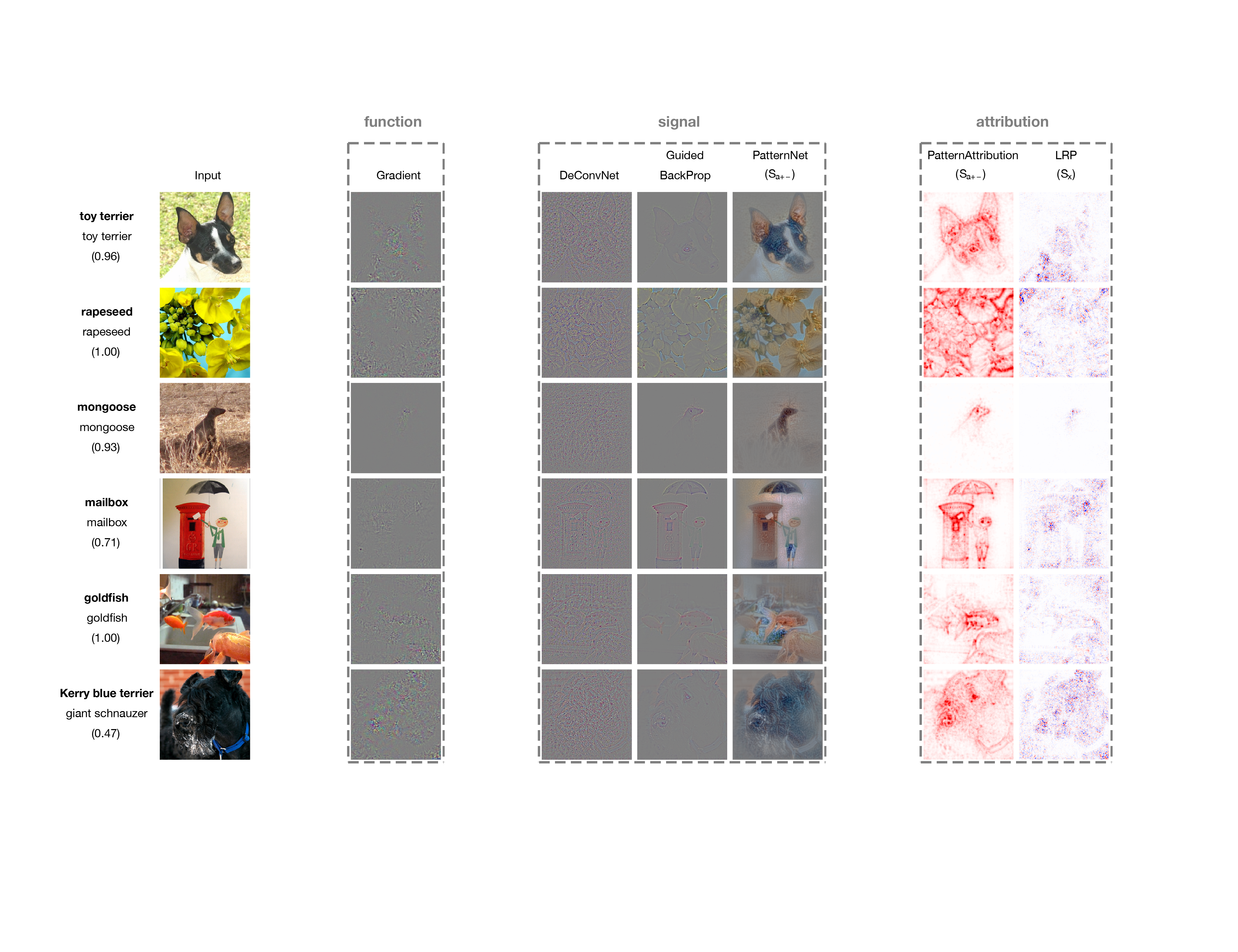}
\caption{Visualization of random images from ImageNet (validation set). In the leftmost shows column the ground truth, the predicted label and the classifier's confidence. Methods should only be compared within their group. PatternNet, Guided Backprop, DeConvNet and the Gradient (saliency map) are back-projections to input space with the original color channels. They are normalized using $x_{norm} = \frac{x}{2\max\left|x\right|}+\frac{1}{2}$ to maximize contrast. LRP and PatternAttribution are heat maps showing pixel-wise contributions. Best viewed in electronic format (zoomed in). The supplementary contains more samples. \label{fig:fig6}}
\end{figure*}

\paragraph{Measuring the quality of signal estimators}
In \fig{fig3} we present the results from the correlation measure $\rho(\bx)$, where higher values are better. 
We use random directions as baseline signal estimators. 
Clearly, this approach removes almost no correlation.
The filter-based estimator $S_{\bw}$ succeeds in removing some of the information in the first layer.
This indicates that the filters are similar to the patterns in this layer.
However, the gradient removes much less information in the higher layers.
Overall, it does not perform much better than the random estimator. {\em This implies that the weights do not correspond to the detected stimulus in a neural network.} Hence the implicit assumptions about the signal made by DeConvNet and Guided BackProp is not valid.
The optimized estimators remove much more of the correlations across the board. 
For convolutional layers, $S_{\ba}$ and $S_{\ba+-}$ perform comparably in all but one layer.
The two component estimator $S_{\ba+-}$ is best in the dense layers.

\paragraph{Image degradation}
The first experiment was a direct measurement of the quality of the signal estimators of individual neurons. The second one is an indirect measurement of the quality, but it considers the whole network. 
We measure how the prediction (after the soft-max) for the initially selected class changes as a function of corrupting more and more patches based on the ordering assigned by the attribution \citep[see][]{Samek2016}. This is also related to the work by~\citet{Zintgraf2017}.
In this experiment, we split the image in non-overlapping patches of 9x9 pixels.
We compute the attribution and sum all the values within a patch.
We sort the patches in decreasing order based on the aggregate heat map value. 
In step $n=1..100$ we replace the first $n$ patches with the their mean per color channel to remove the information in this patch. 
Then, we measure how this influences the classifiers output.
We use the estimators from the previous experiment to obtain the function-signal attribution heat maps for evaluation. 
A steeper decay indicates a better heat map. 

Results are shown in \fig{fig4}.  The baseline, in which the patches are randomly ordered, performs worst. 
The linear optimized estimator $S_{\ba}$ performs quite poorly, followed by the filter-based estimator $S_{\bw}$. 
The trivial signal estimator $S_{\bx}$ performs just slightly better.
However, the two component model $S_{\ba+-}$ leads to the fastest decrease in confidence in the original prediction by a large margin.
Its excellent quantitative performance is also backed up by the visualizations discussed next.

\paragraph{Qualitative evaluation}
In \fig{fig5}, we compare all signal estimators on a single input image.
For the trivial estimator $S_{\bx}$, the signal is by definition the original input image and, thus, includes the distractor.
Therefore, its noisy attribution heat map shows contributions that cancel each other in the neural network.
The $S_{\bw}$ estimator captures some of the structure.
The optimized estimator $S_{\ba}$ results in slightly more structure but struggles on color information and produces dense heat maps. The two component model $S_{\ba+-}$ on the right captures the original input during signal estimation and produces a crisp heat map of the attribution. 

\fig{fig6} shows the visualizations for six randomly selected images from ImageNet. 
PatternNet is able to recover a signal close to the original without having to resort to the inclusion of additional rectifiers in contrast to DeConvNet and Guided BackProp.
We argue that this is due to the fact that the optimization of the pattern allows for capturing the important directions in input space. 
This contrasts with the commonly used methods DeConvNet, Guided BackProp, LRP and DTD, for which the correlation experiment indicates that their implicit signal estimator cannot capture the true signal in the data. 
Overall, the proposed approach produces the most crisp visualization in addition to being measurably better, as shown in the previous section.

\paragraph{Relation to previous methods}
Our method can be thought of as a generalization of the work by \cite{Haufe2014}, making it applicable on deep neural networks. Remarkably, our proposed approach can solve the toy example in section \ref{sec:linearmodels} optimally while none of the previously published methods for deep learning are able to solve this%when relying on their guidelines on how to use the method
~\citep{Bach2015,Montavon2017,Smilkov2017,Mukund2017,Zintgraf2017,Dabkowski2017,Zeiler2014,Springenberg2014}. Our method shares the idea that to explain a model properly one has to learn how to explain it with \citet{Zintgraf2017} and \citet{Dabkowski2017}.  Furthermore, since our approach is after training just as expensive as a single back-propagation step, it can be applied in a real-time context, which is also possible for the work done by \citet{Dabkowski2017} but not for \citet{Zintgraf2017}. 

\section{Conclusion}
Understanding and explaining nonlinear methods is an important challenge in machine learning. Algorithms for visualizing nonlinear models have emerged but theoretical contributions are scarce. 
We have shown that the direction of the model gradient does not necessarily provide an estimate for the signal in the data. 
Instead it reflects the relation between the signal direction and the distracting noise contributions (\fig{fig1}). 
This implies that popular explanation approaches for neural networks (DeConvNet, Guided BackProp, LRP) do not provide the correct explanation, even for a simple linear model.
Our reasoning can be extended to nonlinear models. We have proposed an objective function for neuron-wise explanations. 
This can be optimized to correct the signal visualizations (PatternNet) and the decomposition methods (PatternAttribution) by taking the data distribution into account. 
We have demonstrated that our methods constitute a theoretical, qualitative and quantitative improvement towards understanding deep neural networks.

\subsubsection*{Acknowledgments} This project has received funding from the European Union's Horizon 2020 research and innovation programme under the Marie Sklodowska-Curie grant agreement NO 657679, the BMBF for the Berlin Big Data Center BBDC (01IS14013A), a hardware donation from NVIDIA. We thank Sander Dieleman, Jonas Degraeve, Ira Korshunova, Stefan Chmiela, Malte Esders, Sarah Hooker, Vincent Vanhoucke for their comments to improve this manuscript. We are grateful to Chris Olah and Gregoire Montavon for the valuable discussions.

\bibliographystyle{iclr2018_conference}

\bibliography{iclr2018_conference}

\begin{thebibliography}{27}
\providecommand{\natexlab}[1]{#1}
\providecommand{\url}[1]{\texttt{#1}}
\expandafter\ifx\csname urlstyle\endcsname\relax
  \providecommand{\doi}[1]{doi: #1}\else
  \providecommand{\doi}{doi: \begingroup \urlstyle{rm}\Url}\fi

\bibitem[Bach et~al.(2015)Bach, Binder, Montavon, Klauschen, M{\"u}ller, and
  Samek]{Bach2015}
Sebastian Bach, Alexander Binder, Gr{\'e}goire Montavon, Frederick Klauschen,
  Klaus-Robert M{\"u}ller, and Wojciech Samek.
\newblock On pixel-wise explanations for non-linear classifier decisions by
  layer-wise relevance propagation.
\newblock \emph{PloS one}, 10\penalty0 (7):\penalty0 e0130140, 2015.

\bibitem[Baehrens et~al.(2010)Baehrens, Schroeter, Harmeling, Kawanabe, Hansen,
  and M\"uller]{Baehrens2010}
David Baehrens, Timon Schroeter, Stefan Harmeling, Motoaki Kawanabe, Katja
  Hansen, and Klaus-Robert M\"uller.
\newblock How to explain individual classification decisions.
\newblock \emph{Journal of Machine Learning Research}, 11\penalty0
  (Jun):\penalty0 1803--1831, 2010.

\bibitem[Bergstra et~al.(2010)Bergstra, Breuleux, Bastien, Lamblin, Pascanu,
  Desjardins, Turian, Warde-Farley, and Bengio]{Bergstra2010}
James Bergstra, Olivier Breuleux, Fr{\'e}d{\'e}ric Bastien, Pascal Lamblin,
  Razvan Pascanu, Guillaume Desjardins, Joseph Turian, David Warde-Farley, and
  Yoshua Bengio.
\newblock Theano: A cpu and gpu math compiler in python.
\newblock In \emph{Proc. 9th Python in Science Conf}, pp.\  1--7, 2010.

\bibitem[Dabkowski \& Gal(2017)Dabkowski and Gal]{Dabkowski2017}
Piotr Dabkowski and Yarin Gal.
\newblock Real time image saliency for black box classifiers.
\newblock In \emph{NIPS 2017}, 2017.

\bibitem[Dieleman et~al.(2015)Dieleman, Schl{\"u}ter, Raffel, Olson,
  S{\o}nderby, Nouri, Maturana, Thoma, Battenberg, Kelly, et~al.]{Dieleman2015}
Sander Dieleman, Jan Schl{\"u}ter, Colin Raffel, Eben Olson, S{\o}ren~Kaae
  S{\o}nderby, Daniel Nouri, Daniel Maturana, Martin Thoma, Eric Battenberg,
  Jack Kelly, et~al.
\newblock Lasagne: First release.
\newblock \emph{Zenodo: Geneva, Switzerland}, 2015.

\bibitem[Haufe et~al.(2014)Haufe, Meinecke, G{\"o}rgen, D{\"a}hne, Haynes,
  Blankertz, and Bie{\ss}mann]{Haufe2014}
Stefan Haufe, Frank Meinecke, Kai G{\"o}rgen, Sven D{\"a}hne, John-Dylan
  Haynes, Benjamin Blankertz, and Felix Bie{\ss}mann.
\newblock On the interpretation of weight vectors of linear models in
  multivariate neuroimaging.
\newblock \emph{Neuroimage}, 87:\penalty0 96--110, 2014.

\bibitem[Kindermans et~al.(2016)Kindermans, Sch{\"u}tt, M{\"u}ller, and
  D{\"a}hne]{Kindermans2016}
Pieter-Jan Kindermans, Kristof Sch{\"u}tt, Klaus-Robert M{\"u}ller, and Sven
  D{\"a}hne.
\newblock Investigating the influence of noise and distractors on the
  interpretation of neural networks.
\newblock \emph{arXiv preprint arXiv:1611.07270}, 2016.

\bibitem[Kingma \& Ba(2015)Kingma and Ba]{Kingma2014}
Diederik Kingma and Jimmy Ba.
\newblock Adam: A method for stochastic optimization.
\newblock In \emph{ICLR}, 2015.

\bibitem[Krizhevsky et~al.(2012)Krizhevsky, Sutskever, and
  Hinton]{krizhevsky2012imagenet}
Alex Krizhevsky, Ilya Sutskever, and Geoffrey~E Hinton.
\newblock Imagenet classification with deep convolutional neural networks.
\newblock In \emph{Advances in neural information processing systems}, pp.\
  1097--1105, 2012.

\bibitem[LeCun et~al.(2015)LeCun, Bengio, and Hinton]{Lecun2015}
Yann LeCun, Yoshua Bengio, and Geoffrey Hinton.
\newblock Deep learning.
\newblock \emph{Nature}, 521\penalty0 (7553):\penalty0 436--444, 2015.
\newblock \doi{10.1038/nature14539}.

\bibitem[Mnih et~al.(2015)Mnih, Kavukcuoglu, Silver, Rusu, Veness, Bellemare,
  Graves, Riedmiller, Fidjeland, Ostrovski, et~al.]{mnih2015human}
Volodymyr Mnih, Koray Kavukcuoglu, David Silver, Andrei~A Rusu, Joel Veness,
  Marc~G Bellemare, Alex Graves, Martin Riedmiller, Andreas~K Fidjeland, Georg
  Ostrovski, et~al.
\newblock Human-level control through deep reinforcement learning.
\newblock \emph{Nature}, 518\penalty0 (7540):\penalty0 529--533, 2015.

\bibitem[Montavon et~al.(2017)Montavon, Lapuschkin, Binder, Samek, and
  M{\"u}ller]{Montavon2017}
Gr{\'e}goire Montavon, Sebastian Lapuschkin, Alexander Binder, Wojciech Samek,
  and Klaus-Robert M{\"u}ller.
\newblock Explaining nonlinear classification decisions with deep taylor
  decomposition.
\newblock \emph{Pattern Recognition}, 65:\penalty0 211--222, 2017.

\bibitem[Nguyen et~al.(2016)Nguyen, Dosovitskiy, Yosinski, Brox, and
  Clune]{Nguyen2016}
Anh Nguyen, Alexey Dosovitskiy, Jason Yosinski, Thomas Brox, and Jeff Clune.
\newblock Synthesizing the preferred inputs for neurons in neural networks via
  deep generator networks.
\newblock In \emph{Advances in Neural Information Processing Systems}, pp.\
  3387--3395, 2016.

\bibitem[Russakovsky et~al.(2015)Russakovsky, Deng, Su, Krause, Satheesh, Ma,
  Huang, Karpathy, Khosla, Bernstein, et~al.]{Imagenet2015}
Olga Russakovsky, Jia Deng, Hao Su, Jonathan Krause, Sanjeev Satheesh, Sean Ma,
  Zhiheng Huang, Andrej Karpathy, Aditya Khosla, Michael Bernstein, et~al.
\newblock Imagenet large scale visual recognition challenge.
\newblock \emph{International Journal of Computer Vision}, 115\penalty0
  (3):\penalty0 211--252, 2015.

\bibitem[Samek et~al.(2016)Samek, Binder, Montavon, Lapuschkin, and
  M{\"u}ller]{Samek2016}
Wojciech Samek, Alexander Binder, Gr{\'e}goire Montavon, Sebastian Lapuschkin,
  and Klaus-Robert M{\"u}ller.
\newblock Evaluating the visualization of what a deep neural network has
  learned.
\newblock \emph{IEEE Transactions on Neural Networks and Learning Systems},
  2016.
\newblock \doi{10.1109/TNNLS.2016.2599820}.

\bibitem[Schmidhuber(2015)]{Schmidhuber2015}
J{\"u}rgen Schmidhuber.
\newblock Deep learning in neural networks: An overview.
\newblock \emph{Neural networks}, 61:\penalty0 85--117, 2015.

\bibitem[Shrikumar et~al.(2016)Shrikumar, Greenside, Shcherbina, and
  Kundaje]{LRPGRAD16}
Avanti Shrikumar, Peyton Greenside, Anna Shcherbina, and Anshul Kundaje.
\newblock Not just a black box: Learning important features through propagating
  activation differences.
\newblock \emph{CoRR}, abs/1605.01713, 2016.
\newblock URL \url{http://arxiv.org/abs/1605.01713}.

\bibitem[Silver et~al.(2016)Silver, Huang, Maddison, Guez, Sifre, Van
  Den~Driessche, Schrittwieser, Antonoglou, Panneershelvam, Lanctot,
  et~al.]{Silver2016}
David Silver, Aja Huang, Chris~J Maddison, Arthur Guez, Laurent Sifre, George
  Van Den~Driessche, Julian Schrittwieser, Ioannis Antonoglou, Veda
  Panneershelvam, Marc Lanctot, et~al.
\newblock Mastering the game of go with deep neural networks and tree search.
\newblock \emph{Nature}, 529\penalty0 (7587):\penalty0 484--489, 2016.
\newblock \doi{10.1038/nature16961}.

\bibitem[Simonyan \& Zisserman(2015)Simonyan and Zisserman]{Simonyan2014}
Karen Simonyan and Andrew Zisserman.
\newblock Very deep convolutional networks for large-scale image recognition.
\newblock In \emph{ICLR}, 2015.

\bibitem[Simonyan et~al.(2014)Simonyan, Vedaldi, and Zisserman]{Simonyan2013}
Karen Simonyan, Andrea Vedaldi, and Andrew Zisserman.
\newblock Deep inside convolutional networks: Visualising image classification
  models and saliency maps.
\newblock In \emph{ICLR}, 2014.

\bibitem[Smilkov et~al.(2017)Smilkov, Thorat, Kim, Vi{\'e}gas, and
  Wattenberg]{Smilkov2017}
Daniel Smilkov, Nikhil Thorat, Been Kim, Fernanda Vi{\'e}gas, and Martin
  Wattenberg.
\newblock Smoothgrad: removing noise by adding noise.
\newblock \emph{arXiv preprint arXiv:1706.03825}, 2017.

\bibitem[Springenberg et~al.(2015)Springenberg, Dosovitskiy, Brox, and
  Riedmiller]{Springenberg2014}
Jost~Tobias Springenberg, Alexey Dosovitskiy, Thomas Brox, and Martin
  Riedmiller.
\newblock Striving for simplicity: The all convolutional net.
\newblock In \emph{ICLR}, 2015.

\bibitem[Sundararajan et~al.(2017)Sundararajan, Taly, and Yan]{Mukund2017}
Mukund Sundararajan, Ankur Taly, and Qiqi Yan.
\newblock Axiomatic attribution for deep networks.
\newblock In \emph{ICML 2017}, 2017.

\bibitem[Sutskever et~al.(2014)Sutskever, Vinyals, and Le]{Sutskever2014}
Ilya Sutskever, Oriol Vinyals, and Quoc~V Le.
\newblock Sequence to sequence learning with neural networks.
\newblock In \emph{Advances in neural information processing systems}, pp.\
  3104--3112, 2014.

\bibitem[Yosinski et~al.(2015)Yosinski, Clune, Fuchs, and Lipson]{Yosinski2015}
Jason Yosinski, Jeff Clune, Thomas Fuchs, and Hod Lipson.
\newblock Understanding neural networks through deep visualization.
\newblock In \emph{ICML Workshop on Deep Learning}, 2015.

\bibitem[Zeiler \& Fergus(2014)Zeiler and Fergus]{Zeiler2014}
Matthew~D Zeiler and Rob Fergus.
\newblock Visualizing and understanding convolutional networks.
\newblock In \emph{European Conference on Computer Vision}, pp.\  818--833.
  Springer, 2014.

\bibitem[Zintgraf et~al.(2017)Zintgraf, Cohen, Adel, and Welling]{Zintgraf2017}
Luisa~M Zintgraf, Taco~S Cohen, Tameem Adel, and Max Welling.
\newblock Visualizing deep neural network decisions: Prediction difference
  analysis.
\newblock In \emph{ICLR}, 2017.

\end{thebibliography}

\end{document}